\documentclass[11pt]{article}

\usepackage{amssymb,amsmath,amsthm}
\usepackage{graphicx}
\usepackage{latexsym}
\usepackage[vlined]{algorithm2e}
\usepackage{hyperref}

\usepackage{multicol}
\usepackage[table]{xcolor}
\usepackage{pifont}
\usepackage{stackengine}
\usepackage{float}
\usepackage{subcaption}

\newcommand{\Pot}{\mathcal{P}}
\newcommand{\R}{\mathbb{R}}

\newcommand{\Branch}{\mathrm{Branch}}
\newcommand{\Leaf}{\mathrm{Leaf}}
\newcommand{\set}[1]{\{\, {#1} \,\}}
\newcommand{\spa}[1]{\; #1 \;}
\newcommand{\dotcup}{\spa{\dot\cup}}
\DeclareMathOperator*{\argmin}{arg\,min}
\DeclareMathOperator*{\argmax}{arg\,max}
\newcommand{\prob}[1]{\mathbf P\!\left[\, {#1} \,\right]}

\newcommand{\red}[1]{\emph{\vphantom{g}#1}}
\newcommand{\kl}[2]{D_{\mathrm{KL}}(#1\, \|\, #2)}
\newcommand{\stirling}[2]{\left\{\substack{#1\\#2}\right\}}
\newcommand{\A}{\mathcal{A}}
\newcommand{\B}{\mathcal{B}}

\newcommand{\cmark}{{\color{green!60!black}\ding{51}}}
\newcommand{\xmark}{{\color{red}\ding{55}}}

\SetArgSty{textrm}
\SetKw{Break}{break}

\usepackage{numprint}
\npdecimalsign{.}
\nprounddigits{3}
\usepackage[table]{xcolor}
\usepackage{collcell}
\usepackage{tikz}
\usepackage{adjustbox}
\newcolumntype{H}{>{\collectcell\Heat}r<{\endcollectcell}}
\newcommand\Heat[1]{
    \pgfmathparse{int(100*#1*#1*#1)}
    \edef\HeatCell{\noexpand\cellcolor{black!25!green!\pgfmathresult!white}}%
    \HeatCell$\numprint{#1}$%
}
\newcommand\Heatinv[1]{
    \pgfmathparse{int(100*2^(-#1/100))}
    \edef\HeatCell{\noexpand\cellcolor{white!\pgfmathresult!red}}%
    \HeatCell$#1$%
}
\newcolumntype{R}[2]{%
    >{\adjustbox{angle=#1,lap=\width-(#2)}\bgroup}%
    l%
    <{\egroup}%
}

\let\oldnl\nl
\newcommand{\nonl}{\renewcommand{\nl}{\let\nl\oldnl}}

 \theoremstyle{definition}
 
 \theoremstyle{remark}

 \numberwithin{equation}{section}

\begin{document}

\title{Global Evaluation for Decision Tree Learning}

\author{{\em Fabian Sp\"ah}\thanks{Work done partly while at University of Konstanz; see \href{https://github.com/285714/DecisionTrees}{\tt https://github.com/285714/DecisionTrees}}.\\
Department of Computer Science\\
Boston University, Boston, MA\\
{\tt fspaeh@bu.edu}
\and
{\em Sven Kosub}\\
Department of Computer \& Information Science\\
University of Konstanz, Konstanz, Germany\\
{\tt sven.kosub@uni-konstanz.de}
}

\date{\empty}

\maketitle

\begin{abstract}
  We transfer distances on clusterings to the building process of decision trees,
  and as a consequence extend the classical ID3 algorithm to perform
  modifications based on the global distance of the tree to the ground
  truth---instead of considering single leaves.
  Next, we evaluate this idea in comparison with the original version and
  discuss occurring problems, but also strengths of the global approach.
  On this basis, we finish by identifying other scenarios where global
  evaluations are worthwhile.
\end{abstract}

\maketitle

\section{Decision Tree Learning}

The classification problem in machine learning asks, given some observed
instances with known outcomes (called the labeled training data), to make
predictions on outcomes of unseen instances.

Formally, let $\Omega$ be a universe of instances.
Every $x \in \Omega$ has attributes $x_1, \dots, x_m \in \R$.
Outcomes of instances in the training set $X \subseteq \Omega$, also called
class labels, are given by a map $y : \Omega \to \{1, \dots, k\}$.
One popular choice of a model to train is the decision tree.
We restrict our analysis to binary decision trees; binary trees whose
branches consist of splitting
criteria $c : \Omega \to \{0,1\}$ and whose leaves are class labels
$\{1, \dots, k\}$.
The decision tree models a discrete-valued function.
Every instance $x$ is sorted down the decision tree by evaluating branch
predicates $c$ and recursing into the respective subtree: left if $c(x) = 0$
and right if $c(x) = 1$.
Once a leaf is encountered, its class label is returned.
Decision trees are capable of handling continuous and discrete attributes and
hence a multitude of splitting criteria are commonly used.
Here we restrict ourselves to
binary criteria of the form $c(x) = [x_j \le r]$ for an attribute $j$ and
limit $r \in \R$.

The labels $y$ can also be interpreted as a clustering
$\mathcal Y = (Y_1, \dots, Y_k)$ where every cluster set is defined as
$Y_i = y^{-1}(i) = \set{ x \in X \mid y(x) = i }$.
Hence, we will use $y$ and $\mathcal Y$ interchangeably.
A decision tree $t$ naturally induces a clustering on X as well, denoted by $\mathcal T_t$.
This leads to the idea to utilize distances on clusterings in the
construction process of decision trees.

\subsection{Distances between Clusterings}

To keep the following definitions concise, 
we denote the set of all $k$-clusterings on $X$ as
$\stirling{X}{k} =_{\rm def} \set{ (A_1, \dots, A_k) \mid A_1 \dotcup \cdots \dotcup A_k = X }$
and write
$A \circ \B =_{\rm def} (A \circ B_1, \dots, A \circ B_l) \in \stirling{A \circ X}{l}$
as well as
$\A \circ \B =_{\rm def} (A_1 \circ B_1, \dots, A_i \circ B_j, \dots, A_k \circ B_l) \in
\stirling{X \circ X}{k \cdot l}$
for clusterings $\A \in \stirling{X}{k}, \B \in \stirling{X}{l}$, a set
$A \subseteq X$, and an operation $\circ$ on $\Pot(X)$.
We call a clustering $\A$ \textit{trivial} if $A_i = X$ for an
$i \in \{1, \dots, k\}$.

In the following, let clusterings $\A$ and $\B$ be defined as above.
We understand a real-valued function $f : \stirling{X}{k}^2 \to \R$ as a
\red{distance measure} if $f(\A, \B)$ decreases with $\A$ and $\B$ becoming
more similar, and as a \red{similarity measure} if $f(\A, \B)$ increases.
We always refer to measures as distance measures, sometimes implying that
a similarity measure has to be inverted.
All introduced measures are summarized along with their properties in
Table~\ref{table:dist}, and
a more complete collection of distance measures on clusterings can be found in
\cite{wagner2007}.
The first distance measures we introduce originate in probability theory.

\paragraph{Information-Theoretic Measures}

For subsets $A, B \subseteq X$, define 
\[
  P(A) =_{\rm def} \dfrac{|A|}{|X|} = \prob{x \in A}, \hspace{0.5cm}
  P(A | B) =_{\rm def} \dfrac{P(A \cap B)}{P(B)}
    = \dfrac{|A \cap B|}{|B|}
    = \prob{x \in A \mid x \in B}
\]
where $x$ is a random variable assuming values from $X$ uniformly at random.
Understanding a clustering $\A$ as a probability distribution $P(A_i)$ over
values of $i$ allows
us to transfer probability-theoretic concepts to clusterings:
the (Shannon) \red{entropy associated with a clustering $\A$}
and the \red{expected conditional entropy of a clustering $\A$ given $\B$}
are defined as:
\[
  \begin{array}{rrl}
    \mathrm H(\A) &=_{\rm def}& - \sum_{i=1}^k P(A_i) \log P(A_i) \\[3pt]
    \displaystyle \mathrm H(\A \mid \B) &=_{\rm def}&
    \mathbf E_j\!\left[\, \mathrm H(\A \mid B_j) \,\right] ~=_{\rm def}~
    - \sum_{i=1}^k \sum_{j=1}^l P(A_i \cap B_j) \log P(A_i \mid B_j) 
  \end{array}
\]
Clearly, $\mathrm H(\A | \B) = \mathrm H(\A \cap \B) - \mathrm H(\B)$.

The \red{Kullback-Leibler divergence associated with
$\mathcal C \in \stirling{X}{k}$ and $\mathcal C' \in \stirling{Z}{k}$}
is the relative entropy between $P(C_i)$ and $P(C'_i)$, given as
\[
  \kl{\mathcal C}{\mathcal C'} =_{\rm def}
    \sum_{i = 1}^m P(C_i) \log \frac{P(C_i)}{P(C'_i)} \ .
\]
This definition is not yet suited to compare clusterings as it
does not consider their joint distributions. 
Instead, we measure the divergence of the joint distribution
$P(A_i \cap B_j)$ to its independent counterpart
$P(A_i) P(B_j)$:
The \red{mutual information between two clusterings $\A$ and $\B$} is defined as
\[
  \begin{array}{l}
    \displaystyle \mathrm I(\A, \B) =_{\rm def}
    \kl{\A \cap \B}{\A \times \B} =
    \sum_{i=1}^k \sum_{j=1}^l P(A_i \cap B_j)
      \log \frac{P(A_i \cap B_j)}{P(A_i) P(B_j)} \\[15pt]
    \hspace{4cm}
    = \mathbf E_j\!\left[\, \kl{\A \cap B_j}{ \A} \,\right]
    = \mathrm H(\A) - \mathrm H(\A \mid \B)
   ~ _{\rm def}= \mathrm{Gain}(\A, \B)
  \end{array}
\]
and also known as \red{information gain} in the context of
decision tree learning \cite[p.~58]{mitchell1997}.

The \red{variation of information} (VI) is given as
$\mathrm{VI}(\A, \B) =_{\rm def} \mathrm H(\A \mid \B) + \mathrm H(\B \mid \A)$.
A normalized version
\[
    \mathrm{VI}_{\mathrm n}(\A, \B) =_{\rm def} \dfrac{\mathrm{VI}(\A, \B)}{\mathrm H(\A \cap \B)}
    = 1 - \dfrac{\mathrm{Gain}(\A, \B)}{\mathrm H(\A \cap \B)}
\]
can also take the shape of a normalized information gain
\cite{demantaras1991}.
Previously, the \red{information gain ratio} defined as
\[
  \mathrm{GainRatio}(\A, \B)
  =_{\rm def} \frac{\mathrm{Gain}(\A, \B)}{H(\B)} \ .
\]
was proposed as normalization \cite{quinlan1986}.
Other normalizations take the geometric or arithmetic mean based on the
fact that $\mathrm I(\A, \B) \le \min\{\mathrm H(\A), \mathrm H(\B)\}$.

\paragraph{Gini Impurity}
Let the \red{Gini impurity} \cite{breiman1998} be defined as
\[
  \mathrm{Gini}(\mathcal A, \mathcal B) =_{\rm def}
  \sum_{j=1}^l P(B_j) \left(1 - \sum_{i=1}^k P(A_i | B_j)^2\right) =
  1 - \sum_{j=1}^l \frac{1}{P(B_j)} \sum_{i=1}^k P(A_i \cap B_j)^2 \ .
\]

\paragraph{Jaccard Distance}

For a nonnegative, monotone, and submodular set function $f$ on $X$, let the
distance $D_f : \Pot(X) \to \R$ be defined as
\[
  D_f(\A, \B) =_{\rm def} \sum_{i=1}^k J_f(A_i, B_i)
  = \sum_{i=1}^k \frac{f(A_i \triangle B_i) - f(\emptyset)}{f(A_i \cup B_i)}
\]
where $A \triangle B = (A \cup B) \setminus (A \cap B)$ denotes the symmetric
difference between $A$ and $B$.
$D_f$ is a metric distance function being the sum of
$k$ metric distance functions $J_f$ \cite{kosub2019}.
For the cardinality $f(A) = |A|$, we obtain the \red{extended Jaccard distance}
\[
  D_{|\cdot|}(\A, \B) = k - \sum_{i=1}^k \frac{P(A_i \cap B_i)}{P(A_i \cup B_i)} \ .
\]

\paragraph{Accuracy}
Finally, we define the prediction accuracy of clustering $\B$
for clustering $\A$ as
\[
  \mathrm{acc}(\A, \B) =_{\rm def}
  \frac{1}{|X|} \sum_{i = 1}^k | A_i \cap B_i | =
  \sum_{i=1}^k P(A_i \cap B_i) \ .
\]

\bigskip
\noindent
The following table summarizes the measures together with some properties.

{\renewcommand{\figurename}{Table}
\begin{figure}[H]
 \centering
  \begin{tabular}{ cllcc@{~~~}cc }
    & Measure & & Range &
      {\stackanchor{Permutation}{invariant}}\vspace*{4pt}&
      Metric \\ \hline
    \color{purple}\ding{108} & Information Gain & $\mathrm{Gain}(\A,\B)$ & $[0, \infty]$ &
      \cmark & \xmark \\
    \color{purple}\ding{108} & Gain Ratio & $\mathrm{GainRatio}(\A,\B)$ &  $[0,1]$ &
      \cmark & \xmark \\
    \color{blue}\ding{115} & normalized VI & $\mathrm{VI}_{\mathrm n}(\A,\B)$ & $[0,1]$ &
      \cmark & \cmark \\
    \color{blue}\ding{115} & Gini Impurity & $\mathrm{Gini}(\A,\B)$ & $[0,1]$ &
     \cmark & \xmark \\
    \color{blue}\ding{115} & extended Jaccard & $D_{|\cdot|}(\A,\B)$ & $[0,k]$ &
      \xmark & \cmark \\
    \color{blue}\ding{115} & inverted Accuracy & $1 - \mathrm{acc}(\A,\B)$ & $[0,1]$ &
      \xmark & \cmark \\
  \end{tabular}
    \caption{Overview of distance measures. 
    The first row shows whether a function is
    a distance ({\color{blue}\ding{115}})
    or  a similarity
    ({\color{purple}\ding{108}})
    measure.
   All functions are symmetric, i.e., $\Delta(\A, \B) = \Delta(\B, \A)$.
    A function is called permutation invariant if
    $\Delta(\A, \B) = \Delta((A_{\sigma(1)}, \dots, A_{\sigma(k)}), \B)$ for all
    $\sigma \in S_k$.}

  \label{table:dist}
\end{figure}

\subsection{Local Distance Evaluation: ID3}

The most common algorithm for decision tree learning is the greedy top-down
optimizer ID3 \cite{quinlan1986}, shown in Algorithm~\ref{alg:id3}.

We describe decision trees algebraically using $\Branch(c, t_1, t_2)$ to
denote a tree with splitting criterion $c : \Omega \to \{0,1\}$ and
$t_1$ and $t_2$ as left and right subtrees.
A leaf of class $i \in \{1, \dots, k\}$ is indicated as $\Leaf(i)$ and the
subset of all instances
in the training set $X$ sorted down along the tree into a leaf $v$ are
denoted as $\mathrm{instances}(t, v)$.
A leaf is called pure if $|y(\mathrm{instances}(t, v))| \le 1$
and a splitting criterion $c$ is called valid (regarding a
set of instances $X' = \mathrm{instances}(t, v)$ in a leaf $v$)
if $c|_{X'}$ is non-trivial.
Finally, during tree construction, $\mathrm{exchange}(t, v, t')$ is the tree
resulting from replacing $v$ in $t$ by $t'$.

{\renewcommand{\figurename}{Algorithm}
\begin{figure}[h]
\caption{ID3}\label{alg:id3}
\setlength{\interspacetitleruled}{4pt}%
\setlength{\algotitleheightrule}{-4pt}%
\SetKwInOut{Input}{Input}
\SetKwInOut{Output}{Output}
\begin{algorithm}[H]
  \KwIn{Training data $(X,y)$ and distance measure $\Delta$}
  \KwOut{Decision tree $t$}
  \vspace{2pt}
\end{algorithm}
\begin{algorithm}[H]
  \vspace{-4pt}
  $t \gets \text{empty tree}$ \\
  \While{$\mathcal T_t \not= \mathcal Y$}{
    $(v, X', c) \gets \argmin \{\; \Delta(c|_{X'}, y|_{X'}) \mid
     \textrm{impure leaves $v$ of $t$},\, \textrm{$X' = \mathrm{instances}(t, v)$,}$ \\ 
    \hspace{15.515em} $\textrm{valid splitting critera $c$} \;\}$ \\
    \For{$b \in \{0,1\}$}{
      $i_b \gets \argmax \set{ | (c|_{X'})^{-1}(b) \cap y^{-1}(i) | \mid i \in \{1, \dots, k\} }$ \\
    }
    $t \gets \mathrm{exchange}(t, v, \Branch(c, \Leaf(i_0), \Leaf(i_1)))$ \\
  }
  \Return{$t$}
  \vspace{0pt}
\end{algorithm}
  \vspace{2pt}
\end{figure}}

\noindent
ID3 starts with an empty tree and grows the tree in every iteration
by replacing a leaf with a new branch containing a splitting criterion and
two leaves itself.
The leaf to replace and the splitting criterion $c$ are determined by evaluating
the distance measure $\Delta$ on the instances in every leaf as clustered by $y$
and as clustered by $c$.
The leaf posing an overall minimum on the distance is selected to be split
along the respective splitting criterion.
Splitting leaves that are already pure or splitting along a splitting criterion
which results in an empty leaf is not allowed to ensure that the algorithm
will stop.
Finally, the two new leaves get assigned the label which is most prevalent
amongst their instances.


\paragraph{Bias-Variance Tradeoff}
We briefly consider the following decomposition of the classification
error to identify challenges in decision tree learning.
The expected error of a regressor $\hat f$ for a sample $x \in \Omega$ 
can be written as
\[
  \mathbf E ((y - \hat f(x))^2)
  = \mathrm{Bias}(\hat f(x))^2 + \mathrm{Var}(\hat f(x)) + \sigma^2
  \quad \textrm{where} \quad
  \mathrm{Bias}(\hat f(x)) =_{\rm def} \mathbf E(\hat f(x)) - f(x)
\]
assuming $y = f(x) + \varepsilon$ is a noisy sample of the original distribution
$f$ with $\varepsilon \sim {\cal N}(0, \sigma^2)$.
This motivates the idea to average multiple independent
classifiers $\hat f_1, \dots, \hat f_s$ as we obtain
$\mathrm{Var}(1 / s \cdot \sum_{i=1}^s \hat f_i(x)) = 1 / s \cdot \mathrm{Var}(\hat f(x))$,
assuming equal variance for all regressors.
We train each classifier on a sample drawn
with replacement from the training set.
Training on small samples increases the bias of the final classifier,
but for \textit{strong learners} that adapt arbitrarily well to the training
data (such as decision trees), this is a valuable trade-off.

A similar decomposition of the expected error exists for a classification task
\cite{domingos2000}; instead of averaging multiple classifiers, we take
the mode and obtain a random forest.
Now, consider a dataset that originates from a decision tree.
Here, we also have a concept of bias and variance in tree structure:
How similar is a tree created by the learning algorithm to the original on
average, and how much deviate learned trees from each other.
These ideas correspond closely to the notions of bias
and variance defined above.

\medskip

Another common way to balance this equation is by pruning.
Techniques such as early stopping limit the tree height or the minimum
number of instances per leaf.
This increases the bias but on the other hand, prevents the tree from overfitting
and thus lowers its variance.

\subsection{Global Distance Evaluation}

We now want to explore a strategy that still grows the tree by performing local
modifications, but decides on them based on evaluations of a distance measure
on the whole training set, not only on the instances in a single leaf
(see Algorithm~\ref{alg:globaltree}). We call this \red{global evaluation}.

\renewcommand{\figurename}{Algorithm}
\begin{figure}[h]
\caption{ID3 with Global Evaluation}
\label{alg:globaltree}
\setlength{\interspacetitleruled}{2pt}%
\setlength{\algotitleheightrule}{0pt}%
\begin{algorithm}[H]
  \KwIn{Training data $(X,y)$ and distance measure $\Delta$}
  \KwOut{Decision tree $t$}
  \vspace{2pt}
\end{algorithm}
\begin{algorithm}[H]
  $t \gets \text{empty tree}$ \\
  \While{$\mathcal T_t \not= \mathcal Y$}{
    $t' \gets \argmin \{\; \Delta(\mathcal T_{t'}, \mathcal Y) \mid
     \textrm{impure leaves $v$ of $t$},\, \textrm{valid splitting critera $c$},$ \\ 
    \hspace{10.7em} $\textrm{class labels $i_0, i_1 \in \{1, \dots, k\}$},$ \\ 
    \hspace{10.7em}  $t' = \mathrm{exchange}(t, v, \Branch(c, \Leaf(i_0), \Leaf(i_1))) \;\}$ \\
    $t \gets t'$
 }
  \Return{$t$}
  \vspace{0pt}
\end{algorithm}
\end{figure}}

In classical ID3 we evaluate our distance measures between clusterings
on the instances in the current leaf, namely the 2-clustering given by the
splitting criterion and the clustering given by the ground truth $y$.
To evaluate globally, we have to decide on the labels
$i_0, i_1$ assigned to both sides of the split beforehand.
Algorithm~\ref{alg:globaltree} evaluates the distance measure on all combinations of
class labels and decides for the one inducing the minimum distance.
A more efficient strategy is to assign the most prevalent class.


\section{Evaluation}
\label{sec:evaluation}

\begin{figure}[p]
  \def \height {2.5cm}\relax
  \def \width {13.2cm}\relax
\def \secwidth {2.8cm}\relax
  \centering
  \makebox[0pt][c]{\begin{minipage}[c]{0.7\paperwidth}
  \footnotesize 
  \begin{tabular}{p{1pt}p{1pt}p{1pt}p{1pt}}
    &&& \begin{tabular}{p{\secwidth}p{\secwidth}p{\secwidth}p{\secwidth}} \centering ~~~Information Gain & \centering Gini Impurity & \centering extended Jaccard & \centering normalized VI \end{tabular} \\
    & \rotatebox{90}{Iris} & \rotatebox{90}{(150s, 4f, 3c)} &
    \begin{subfigure}[t]{\width}
      \includegraphics[width=\textwidth,height=\height]{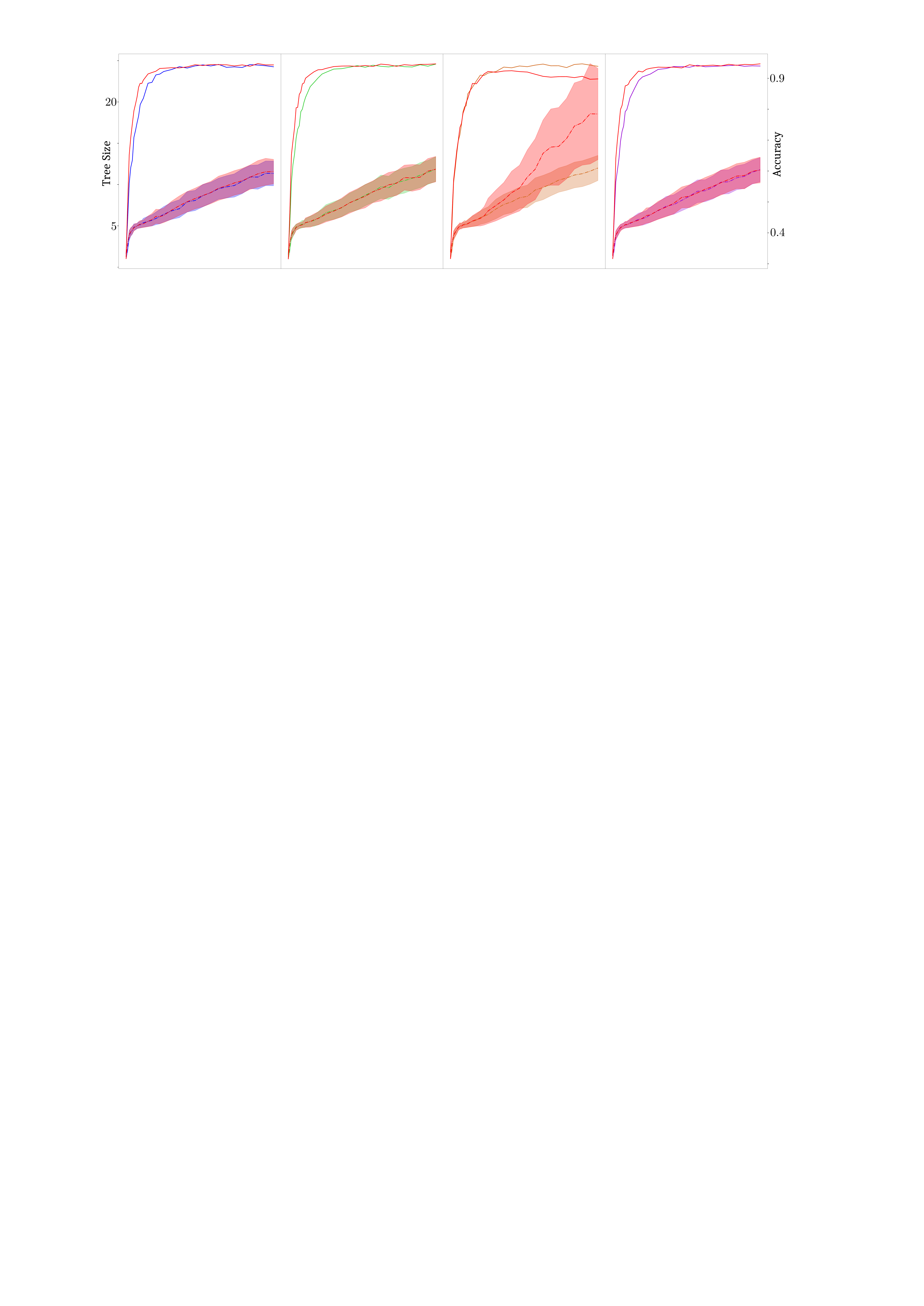}
    \end{subfigure} \\
    & \rotatebox{90}{Cardiotocography} & \rotatebox{90}{(2126s, 13f, 3c)} &
    \begin{subfigure}[t]{\width}
      \hspace{-5.45pt}
      \includegraphics[width=1.0045\textwidth,height=\height]{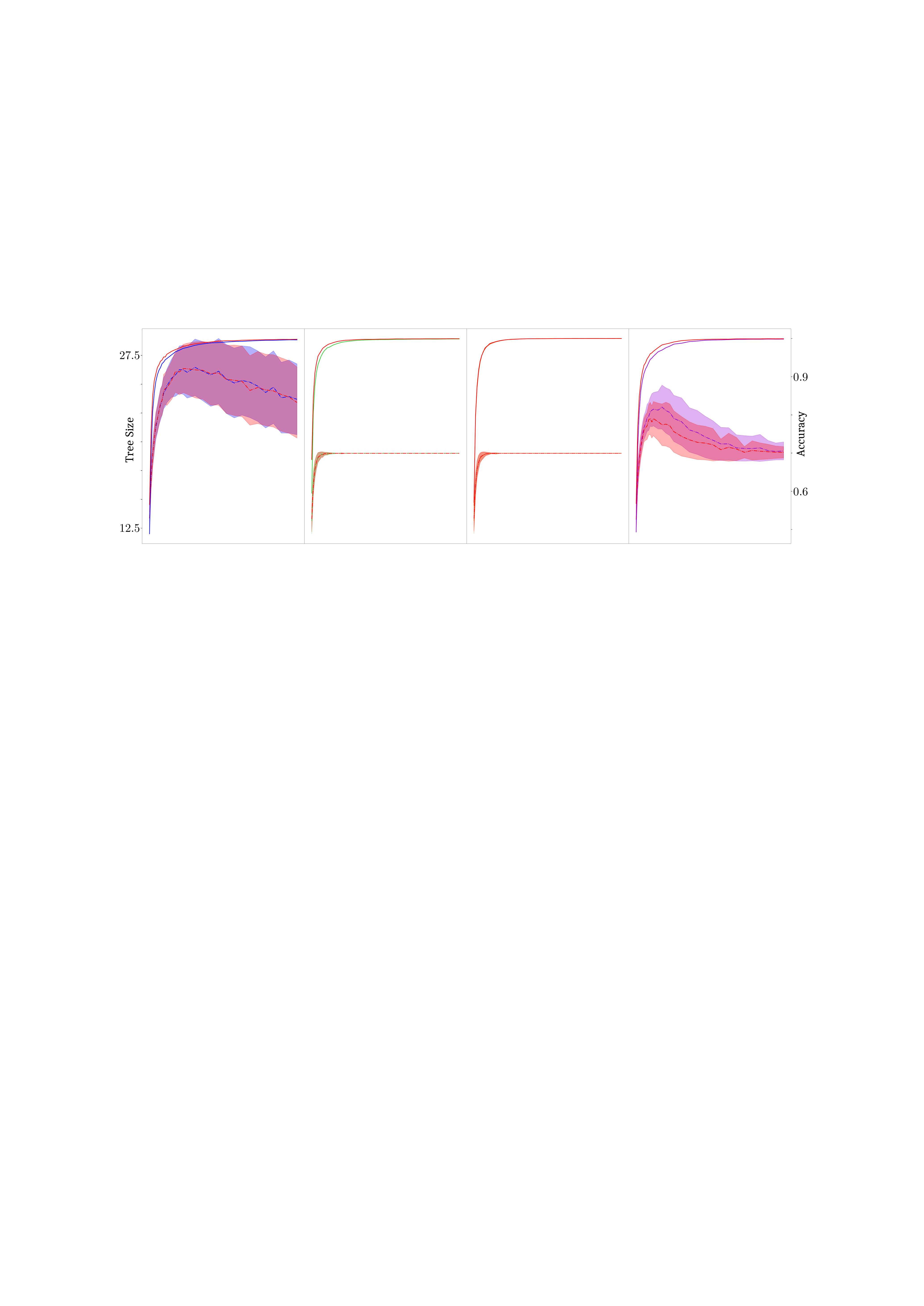}
    \end{subfigure} \\

    \rotatebox{90}{Natural Data} & \rotatebox{90}{Wine} & \rotatebox{90}{(178s, 13f, 3c)} &
    \begin{subfigure}[t]{\width}
     \hspace{-3.2pt}
      \includegraphics[width=0.9995\textwidth,height=\height]{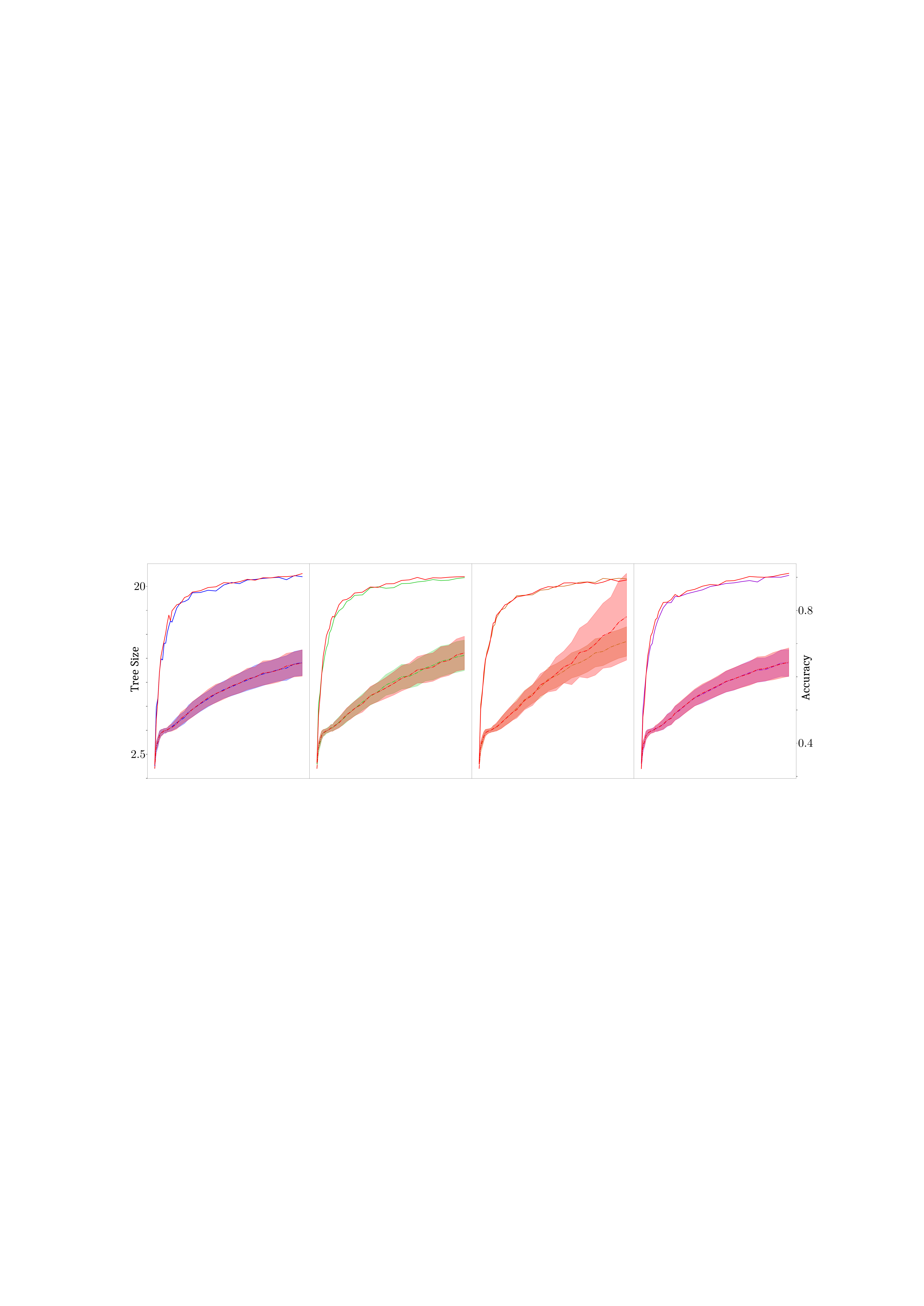}
    \end{subfigure} \\

   & \rotatebox{90}{Gaussian Blobs} & \rotatebox{90}{(2000s, 3f, 3c)} &
    \begin{subfigure}[t]{\width}
      \hspace{-7.15pt}
      \includegraphics[width=1.0075\textwidth,height=\height]{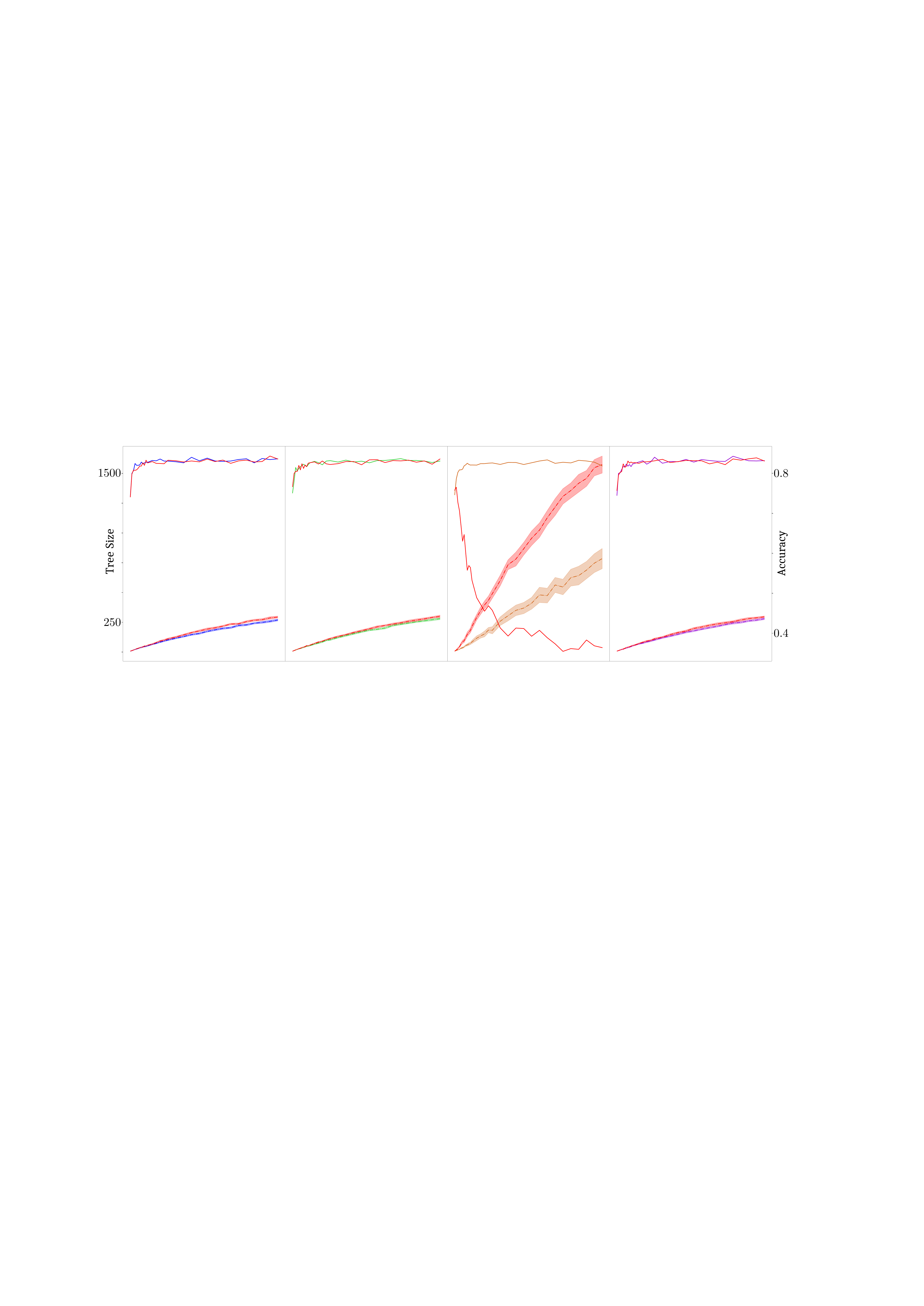}
    \end{subfigure} \\

    & \rotatebox{90}{Monks} & \rotatebox{90}{(556s, 6f, 2c)} &
    \begin{subfigure}[t]{\width}
      \hspace{-3.5pt}
      \includegraphics[width=\textwidth,height=\height]{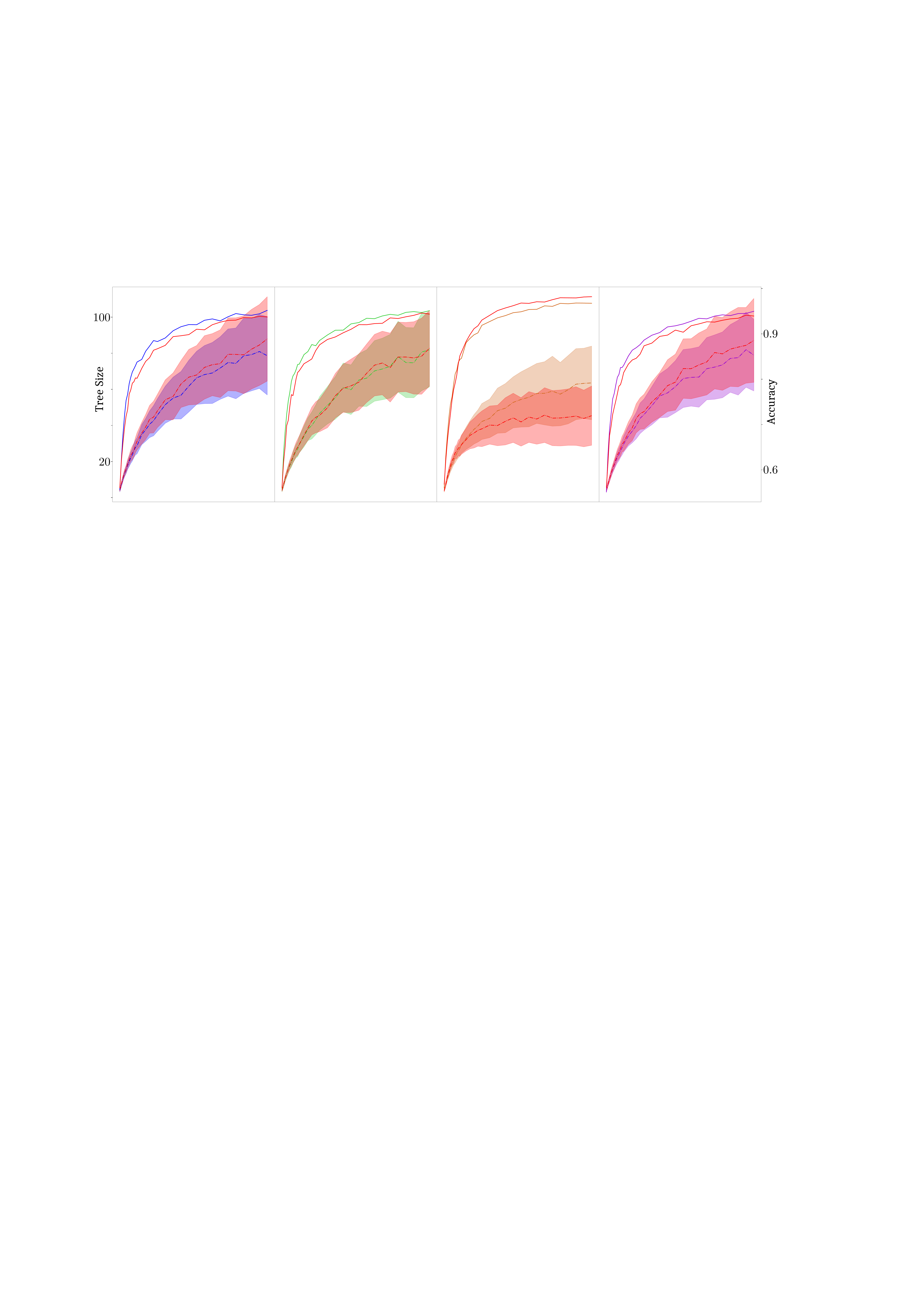}
    \end{subfigure} \\

    & \rotatebox{90}{Monks 2} & \rotatebox{90}{(602s, 6f, 2c)} &
    \begin{subfigure}[t]{\width}
      \hspace{-3.25pt}
      \includegraphics[width=1.0105\textwidth,height=\height]{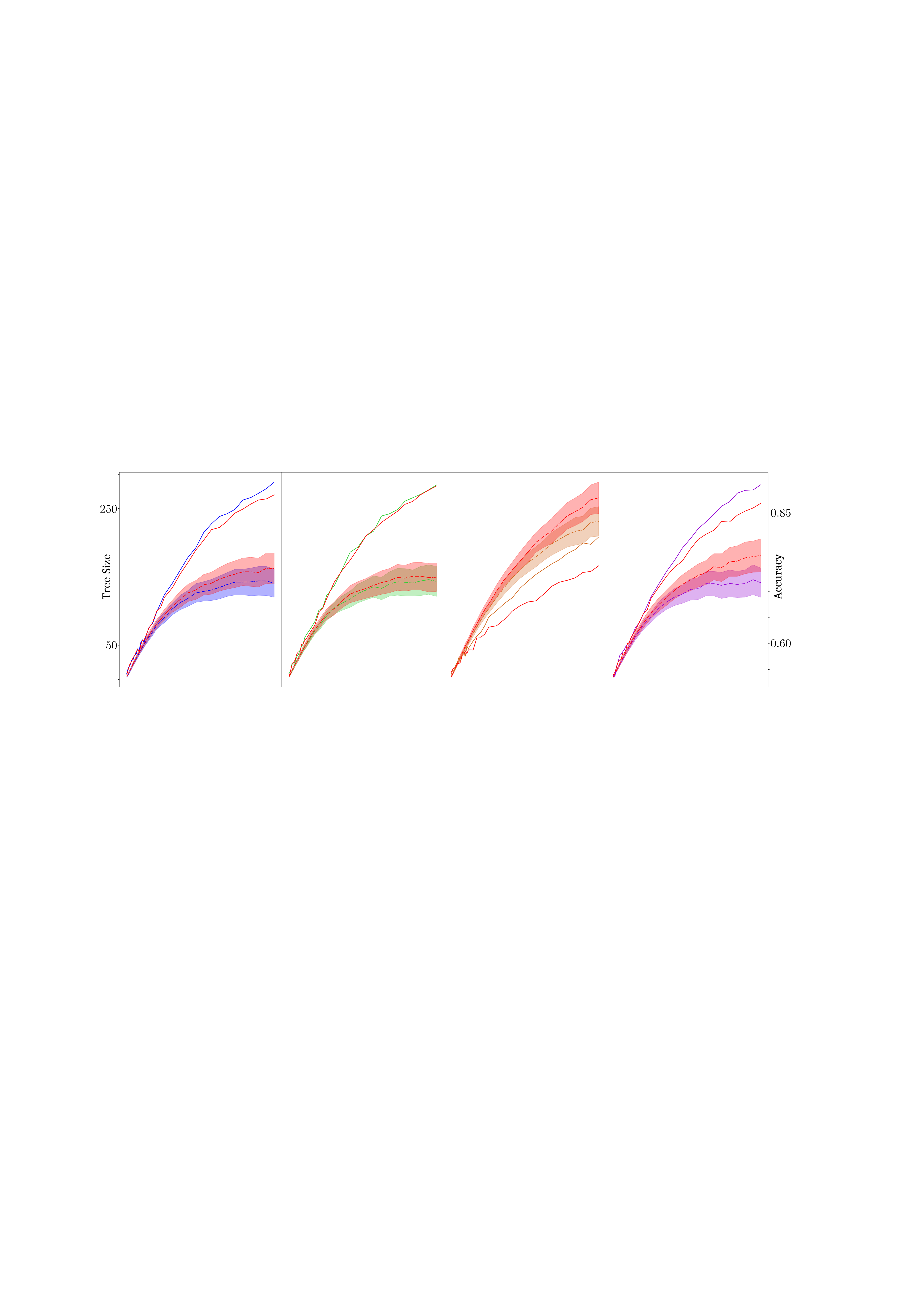}
    \end{subfigure} \\

    \rotatebox{90}{Artificial Data} & \rotatebox{90}{Monks 3} & \rotatebox{90}{(554s, 6f, 2c)} &
    \begin{subfigure}[t]{\width}
      \hspace{-3.3pt}
      \includegraphics[width=1.01\textwidth,height=\height]{Figures/tree-size-and-accuracy_backup_12-09-2019--10-45-44.pdf}
    \end{subfigure} \\

  \end{tabular}
  \caption{The plot shows test accuracies and tree sizes of decision trees. Each
  column corresponds to one distance measure, each row to one dataset. Every cell
  shows statistics for both the global and the local version of ID3 (where
  the global version is always shown in red).
  The continuous line corresponds to the accuracy of the tree as evaluated
  on a test set consisting of 10\% of the data.
  On the x-axis is the size of the train set drawn without replacement from the
  remaining 90\% of the data.
  The tree size is indicated by the dashed line and surrounded by its
  confidence interval.
  Every training process was repeated 500 times to ensure significance.
  }\label{fig:accts}
  \end{minipage}}
\end{figure}

After introducing the most common distance measures and the two versions of ID3,
we turn towards the evaluation of their performance.
We use the Monks, Wine, Cardiotocography, and Iris datasets from the UCI
Machine Learning Repository \cite{uci2019}
and isotropic Gaussian blobs.

\subsection{Test Accuracy and Tree Size}

Typical evaluation criteria after training of a classification algorithm are
train and test accuracy.
Both values can tell us whether our classificator is over- or underfitted.
As we ignore pruning for now, which means the trained trees 
always fit perfectly to the training data, we can only evaluate how overfitted
the tree is through the test accuracy.

Now, consider a scenario where a decision tree is trained on data directly sampled
from another tree. Of course, a \textit{good} decision tree would not only
achieve high accuracy but also resemble the original tree as closely as
possible.
This makes sense as such an estimator achieves low bias on the tree
structure, and hence generalizes better on unseen data.
Also, decision trees are sought to be humanly interpretable for gaining
insights into the given data.
If the data is in fact not sampled from a tree, it is still worthwhile to
uncover an approximate hidden tree structure, though it is unclear what this
looks like.
As such, we always consider the number of nodes as a measure of tree size
in our evaluation.

\paragraph{Findings}
We can see these properties in Figure~\ref{fig:accts} for
several artificial and natural datasets along with the distance measures that were generating
the best results for both local and global evaluation, namely
information gain, Gini impurity, the extended Jaccard distance, and the
normalized variation of information.
The trees were trained on increasing sample sizes, to simulate scenarios where
the structure in the data is over- or under-represented by a sample.

Generally, there is a discrepancy between artificial and natural data.
Global and local evaluation perform similarly on \textit{natural data}.
Only trees grown with the extended Jaccard distance on the Iris dataset overfit the
data which results in high tree size and variance thereof.
We also observe that the global evaluation outperforms local evaluation
on small training sets.

The situation changes when looking at the four \textit{artificial datasets}.
Only when using Gini impurity do both classifiers perform equally well,
otherwise, local evaluation achieves higher accuracy and lower tree size.
However, using the extended Jaccard distance in the Monks 1 dataset is superior to all
other distance measures, in particular when evaluated globally.
Though on Gaussian Blobs and Monks 3, the extended Jaccard distance
fails if evaluated globally.

Overall, we note that low tree size goes in hand with good accuracy.
A tree of lower complexity does not necessarily represent the data
better \cite[p.~65]{mitchell1997}, but judging from our evaluation, this is
empirically correct.
In the next section, we will look into minimal representations in more detail.

\subsection{Tree Size on Artificial Data}

\begin{figure}[h]
  \centering
  \begin{tabular}{*{3}{p{0.3\textwidth}}}
    \centering Information Gain & \centering Gain Ratio & \centering normalized VI
  \end{tabular}
  \includegraphics[width=\textwidth]{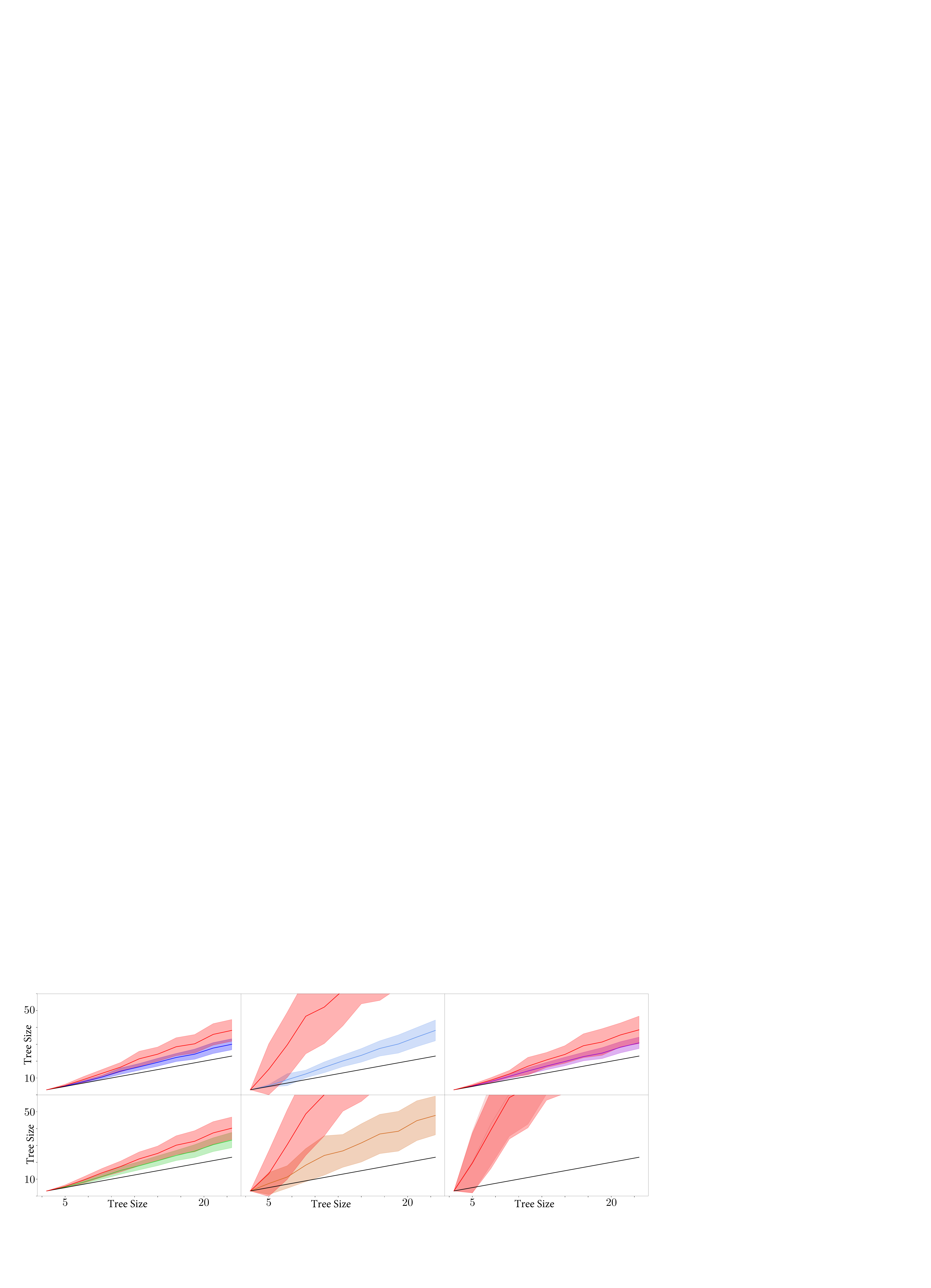}
  \begin{tabular}{*{3}{p{0.3\textwidth}}}
    \centering Gini Impurity & \centering extended Jaccard & \centering Accuracy \\
  \end{tabular}
  \vspace{-0.5cm}
  \caption{Mean tree size and standard deviation of decision trees created by
  global and local evaluation.
  Again, the globally learned trees appear in red.
  The y-axis (and the black line) shows the minimum tree size necessary to
  represent a certain sample of 500 vectors with 6 features of 4 classes.
  Experiments for every tree size were repeated 200 times.
  }
  \label{fig:mts}
\end{figure}

\noindent
Given a sample $(X,y)$, a tree is \red{consistent} if it classifies all
data points in $X$ correctly and
\red{minimal} if its number of nodes is minimum among all consistent trees.
In Figure~\ref{fig:mts}, we evaluate the tree size of classifiers with respect
to the size of the tree the training data was sampled from, by showing the
number of nodes in a tree.
Trees and samples are randomly generated, but discarded if one classifier
finds a consistent tree that is smaller than the original.
This ensures that the original trees are minimal for the produced samples and
we get a better estimate of how close a classifier is
to the original.

Global evaluation generates larger trees for every distance measure.
Local evaluation of information gain, normalized variation of information,
and Gini impurity generates the smallest trees and global evaluation of these
distance measures is not much worse.
The global versions of all other algorithms create unnecessarily large trees.
However, the given data is heavily biased and we already saw in
Figure~\ref{fig:accts} that
an algorithm's performance depends on the distribution it is sampled from,
and in particular whether that is natural or artificial.



\section{Problems}

The above results may seem contradictory.
We chose local manipulations to a tree $t$ to decrease the distance
between $\mathcal T_t$ and the ground truth $\mathcal Y$ in a greedy fashion.
In the end, we want the distance to vanish, i.e., achieve equal clusters.
Should global evaluation not result in faster convergence of $\mathcal T_t$
towards $\mathcal Y$ as local evaluation, since it considers all instances
in every step?
Also, should accuracy as a simple measure of deviance from equality not deliver
the best results?
We will try to find reasons in the following.

\subsection{Minimal Trees}

Although finding minimal trees is NP-complete \cite{hancock1996}, we can find
such a tree using the greedy ID3 algorithm with the following distance measure.
Define
\[
  \begin{array}{l}
    \mathrm{mts}(\mathcal C) =_{\rm def} \min \set{\mathrm{size}(t) \mid
      \textrm{decision tree $t$ with $\mathcal T_t = \mathcal C$}}
    \quad \textrm{and} \\ \hspace{18em}
    \Delta_{\textrm{mts}}(\mathcal A, \mathcal B) =_{\rm def} \sum_{i=1}^k
      \mathrm{mts}(A_i \cap \mathcal B)
  \end{array}
\]
where $\mathrm{size}(t)$ is the number of nodes in $t$.
Note that unlike previous distance measures, this is an internal measure looking
into attributes of the vectors $x \in X$, rather than solely class labels.
Using this distance in the locally evaluating version of ID3 results in a
minimal tree.
Growing the tree according to global evaluations of any distance measure
inevitably fails to produce minimal trees as the clustering $\mathcal T_t$ is
agnostic of the tree structure.

We can understand $\mathrm{mts}(\mathcal Y)$ as a measure of complexity
in the clustering $\mathcal Y$ and, for example, the entropy $\mathrm H(\mathcal Y)$
as an efficiently-computable estimator thereof.
The above construction suggests that using a distance measure
$\Delta((A_1, \dots, A_k), \mathcal Y)$ which is an
aggregate of the approximate clustering complexities in $A_i \cap \mathcal Y$ leads to
short trees.

\subsection{Redundant Splitting}

A redundant split occurs if the generated tree contains a subtree of the form
\[
  \Branch(c_1, \Leaf(i), \Branch(c_2, \Leaf(i), t))
\]
where $c_1(x) = [x_j \le r_1]$ and $c_2(x) = [x_j \le r_2]$ for a common
feature $j \in \{1, \dots, m\}$, class label $i \in \{1, \dots, k\}$, and
subtree $t$.
The same classification can be achieved by a smaller tree
$\Branch(c, \Leaf(i), t)$ with $c(x) = [x_j \le \max\{r_1, r_2\}]$.
Redundant splits can occur during decision tree building if every split
trough the instances of a leaf increases the distance to the ground truth.
ID3 will then choose to split along a plane having fewest instances on one side,
thus altering the current clustering $\mathcal T_t$---and
decreasing the distance---only slightly.

\begin{figure}[H]
  \begin{multicols}{2}
    \begin{subfigure}[t]{0.5\textwidth}
      \centering
      \scalebox{0.7}{\begin{tikzpicture}
        \fill[red!10] (5,2) rectangle (6,3);
        \fill[blue!10] (6,2) rectangle (7,3);
        \node[circle,fill=red,draw=black] at (5.5,3.5) {};
        \node[circle,fill=blue,draw=black] at (6.5,3.5) {};
        \draw[help lines, color=gray!60, dashed] (5,2) grid (7,4);
        \node at (4.5,2.5) {$\A$};
        \node at (4.5,3.5) {$\B$};
       \node at (2.5,3.5) {current leaf};
        \begin{scope}[even odd rule]
          \clip {(0,3) rectangle (1,4)} {(-0.5,2.5) rectangle (1.5,4.5)};
          \draw[line width=6pt, yellow!40, rounded corners=1pt] (0,3)--(1,3)--(1,4)--(0,4)--cycle;
        \end{scope}
        \draw[thick] (0,3)--(1,3)--(1,4)--(0,4)--cycle;

       \draw[help lines, color=gray!60, dashed] (0,0) grid (7,1);
        \draw[->,thick] (-0.5,0)--(7.5,0) node[right] {$x_1$};
        \begin{scope}[even odd rule]
          \clip {(0,0) rectangle (7,1)} {(-0.5,-0.5) rectangle (7.5,1.5)};
          \draw[line width=6pt, yellow!40, rounded corners=1pt] (0,1)--(7,1)--(7,0)--(0,0)--cycle;
        \end{scope}
        \draw[thick] (0,0)--(0,1)--(7,1)--(7,0)--cycle;
        \foreach \i in {1,...,7} {
         \def \col {red}
          \ifthenelse{\i = 4}{ \def \col {blue} }{}
          \node[circle,fill=\col,draw=black] at (\i-0.5,0.5) {};
        }
        \draw[thick,gray] (4,-0.5) node[below] {$c_1$} --(4,1.5);
        \draw[thick,gray] (1,-0.5) node[below] {$c_2$} --(1,1.5);
      \end{tikzpicture}}
      \vspace{-0.2cm}
      \captionsetup{justification=centering}
      \caption{Local evaluation}
    \end{subfigure}
    \vspace{0.5cm}

    \setcounter{subfigure}{2}
    \begin{subfigure}[t]{0.5\textwidth}
      \centering
{\footnotesize
      \begin{tabular}{ lcc }
        Measure & \textsf{\textbf{(L)}} & \textsf{\textbf{(G)}} \\ \hline
        information gain & \cmark & \xmark \\
        gain ratio & \cmark & \xmark \\
        normalized VI & \cmark & \xmark \\
        Gini impurity & \cmark & \xmark \\
        extended jaccard & \xmark & \xmark \\
        inverted accuracy & \xmark & \xmark \\
     \end{tabular}
     }
     \captionsetup{justification=centering}
      \caption{Efficient splits}
    \end{subfigure} \\

    \setcounter{subfigure}{1}
    \begin{subfigure}[t]{0.5\textwidth}
      \centering
      \scalebox{0.7}{\begin{tikzpicture}
        \fill[red!10] (0,1) rectangle (7,7);
        \fill[blue!10] (4,1) rectangle (7,4);
        \fill[blue!10] (3,4) rectangle (7,7);
        \draw[help lines, color=gray!60, dashed] (0,0) grid (7,7);
        \draw[->,thick] (-0.5,0)--(7.5,0) node[right] {$x_1$};
        \draw[->,thick] (0,-0.5)--(0,7.5) node[above] {$x_2$};
        \draw[thin] (0,7)--(7,7)--(7,1);
        \draw[thin] (4,1)--(4,4)--(3,4)--(3,7);
       \begin{scope}[even odd rule]
        \clip {(0,0) rectangle (7,1)} {(-0.5,-0.5) rectangle (7.5,1.5)};
          \draw[line width=6pt, yellow!40, rounded corners=1pt] (0,1)--(7,1)--(7,0)--(0,0)--cycle;
        \end{scope}
        \draw[thick] (0,1)--(7,1)--(7,0)--(0,0)--cycle;
        \foreach \i in {1,...,7} {
          \foreach \j in {1,...,7} {
            \def \col {red}
            \ifthenelse{\j = 1 \AND \i = 4}{ \def \col {blue} }{}
           \ifthenelse{\j > 1 \AND \i > 4}{ \def \col {blue} }{}
            \ifthenelse{\j > 4 \AND \i > 3}{ \def \col {blue} }{}
            \node[circle,fill=\col,draw=black] at (\i-0.5,\j-0.5) {};
          }
        }
        \draw[thick,gray] (4,-0.5) node[below] {$c_1$} --(4,1.5);
        \draw[thick,gray] (1,-0.5) node[below] {$c_2$} --(1,1.5);
      \end{tikzpicture}}
      \vspace{-0.2cm}
      \captionsetup{justification=centering}
     \caption{Global evaluation}
    \end{subfigure} \\
  \end{multicols}

  \caption{Classification surface for local and global optimization. The colored
  dots represent the ground truth $\mathcal Y$ whereas the background-color
  represents the clustering induced by the decision tree being built.
  Introducing a leaf that splits at $c_2$ leads to a redundant split while
  splitting at $c_1$ is the most efficient split, making fastest progress
  towards a consistent tree; (c) shows
  which distance measure induces an efficient split, in both the local
  \textsf{\textbf{(L)}} and global \textsf{\textbf{(G)}}
  setting.}
  \label{fig:redundant}
\end{figure}
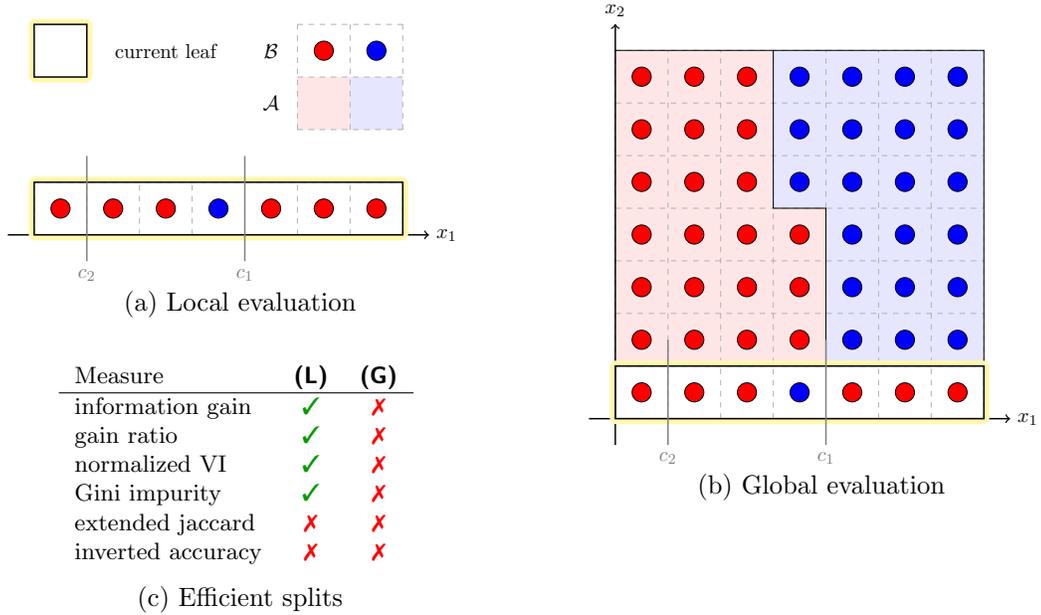

Figure~\ref{fig:redundant} shows two settings, a local and a global one, in
which redundant splits can occur.
It turns out that all distance measures are susceptible to
redundant splitting when evaluated globally.
Using the extended Jaccard distance and inverted accuracy also exhibit
behavior this when evaluated locally.

\section{Pruning and \textit{Glocal} Evaluation}

In the previous sections, we discussed that global evaluation generally
creates trees with lower test accuracy and higher tree size than
local evaluation. This is due to a higher bias of the ID3 algorithm using
global evaluation which overgrows most trees.
However, in a scenario with small sample size or an overly noisy sample,
we benefit from an increased bias.
The former was already apparent in Section~\ref{sec:evaluation}, so we will
discuss the latter in this final section.
Since both local and global evaluation fit arbitrarily well to the
training data, we have to employ pruning strategies to prevent overfitting.

\begin{figure}[H]
  \begin{subfigure}[t]{0.47\textwidth}
    \centering
    \includegraphics[width=\textwidth]{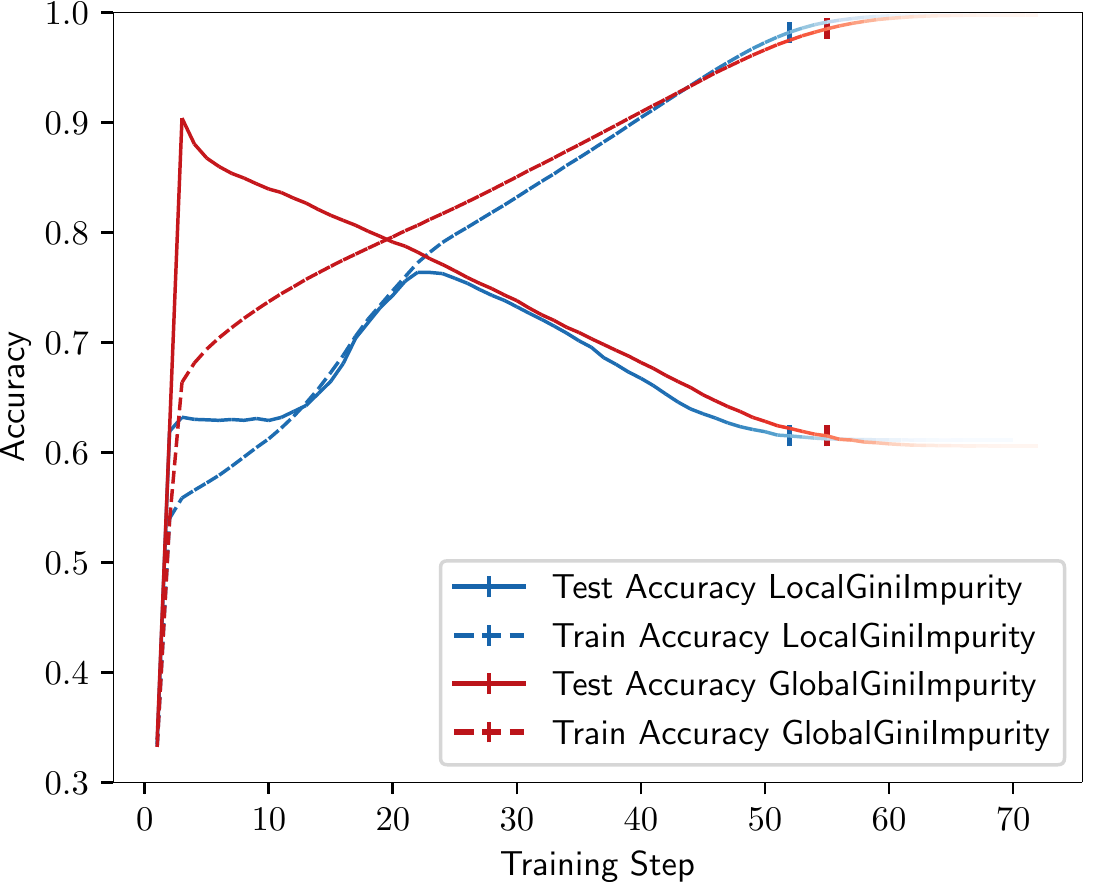}
    \captionsetup{justification=centering}
    \caption{No pruning
      (Gini impurity)}
    \label{fig:pruning:no-pruning}
  \end{subfigure}  ~~~~
  \begin{subfigure}[t]{0.47\textwidth}
    \centering
    \includegraphics[width=\textwidth]{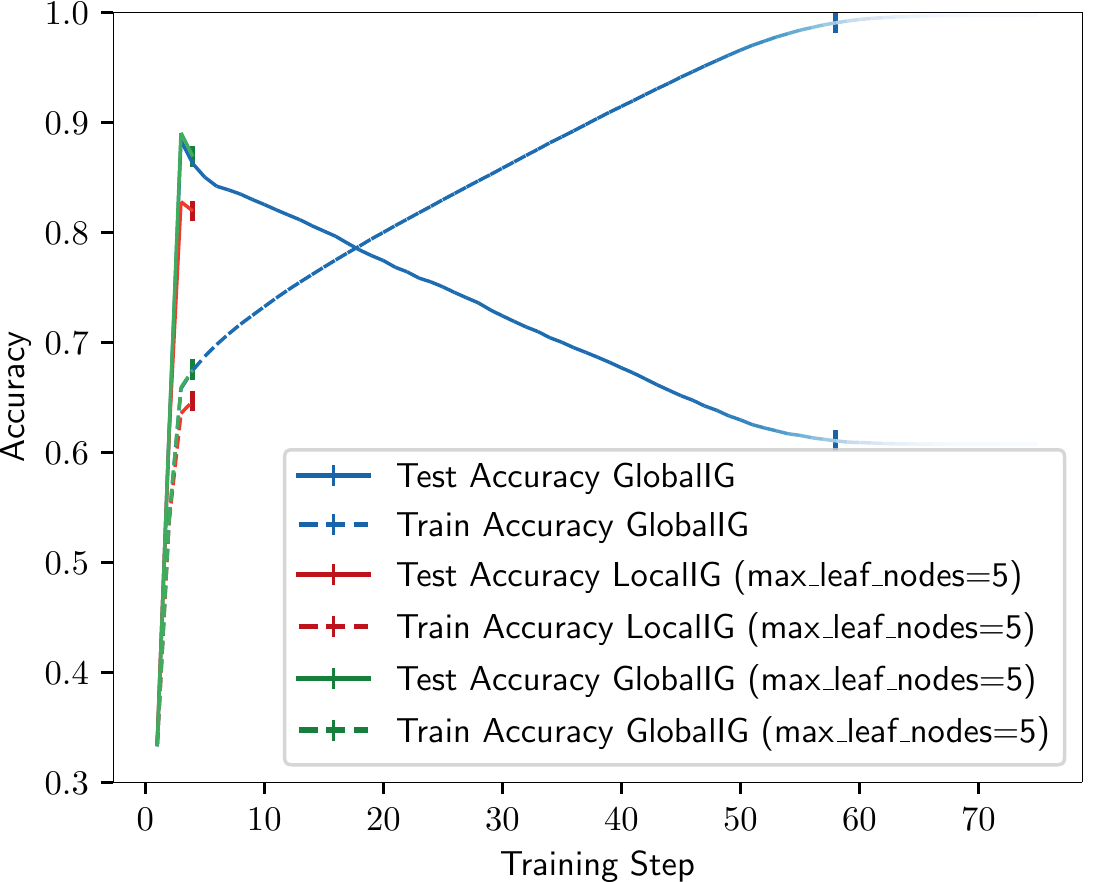}
    \captionsetup{justification=centering}
    \caption{Limit on the number of nodes
      (information gain)}
  \end{subfigure} \\[0.4cm]

  \begin{subfigure}[t]{0.47\textwidth}
    \centering
    \includegraphics[width=\textwidth]{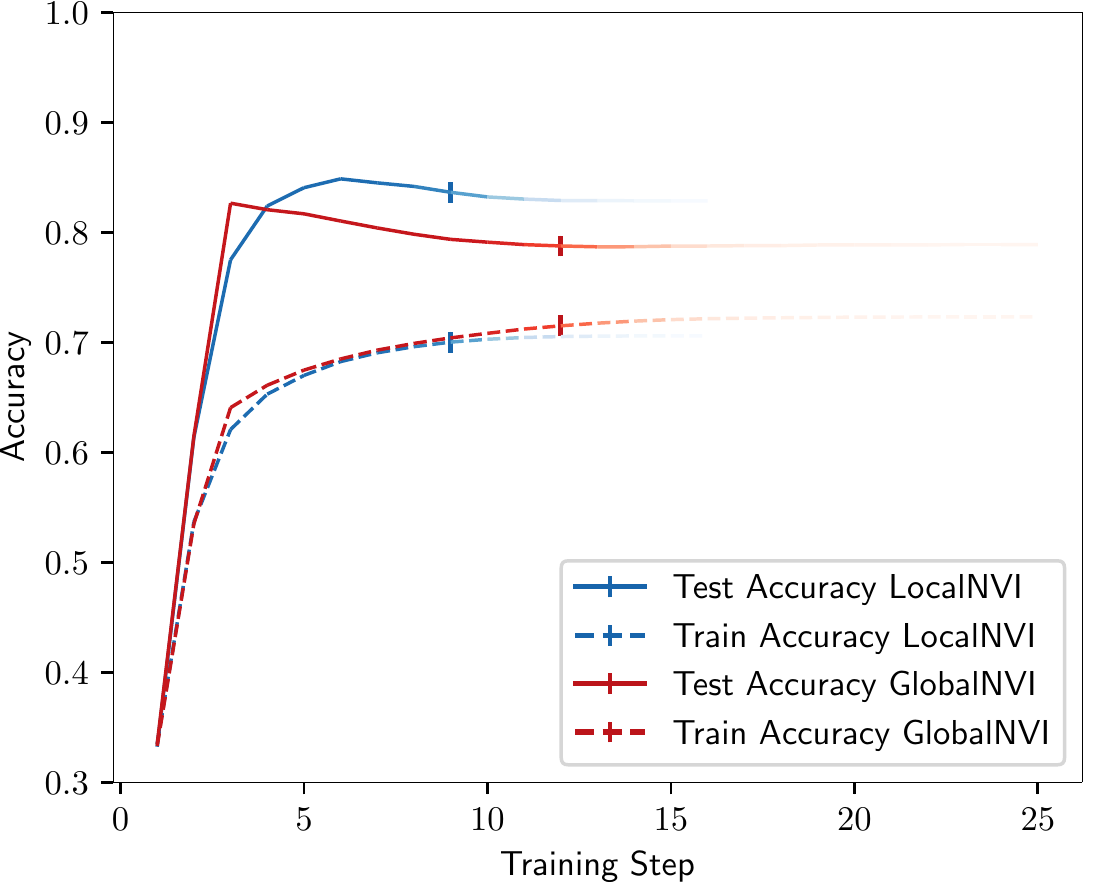}
    \captionsetup{justification=centering}
    \caption{Requiring a minimum of instances in each branch
      (normalized variation of information)}
  \end{subfigure} ~~~~
  \begin{subfigure}[t]{0.47\textwidth}
    \centering
    \includegraphics[width=\textwidth]{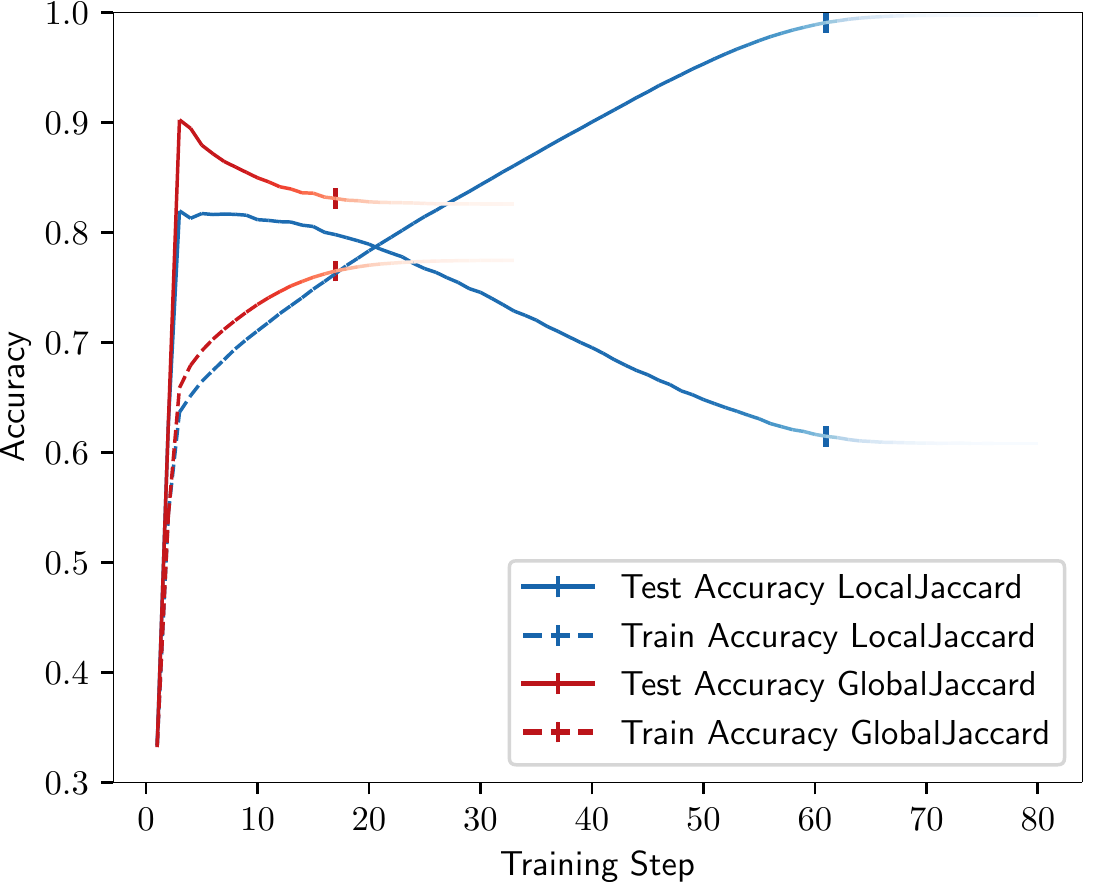}
    \captionsetup{justification=centering}
    \caption{No distance decay
      (extended Jaccard distance)}
  \end{subfigure}
  \caption{Classification accuracy during training on noisy data using different
  pruning techniques.
  The data was obtained by assigning random classes for 50\% of the instances
  of the Iris dataset. The dashed line indicates the train accuracy on the
  noisy data, while the solid line indicates the test accuracy on the original
  data.
  The training process was repeated 1000 times and averaged. The opacity shows
  the percentage of unfinished trees in each step and the horizontal tick
  marks where the majority of trees finished training.}
  \label{fig:pruning}
\end{figure}

\newpage 
\noindent
In Figure~\ref{fig:pruning}, we see average train and test accuracies of a tree
during its construction, as well as the average final train and test accuracy.
We discuss different pruning strategies in the following.

\begin{enumerate}
  \item \label{pruning:a} \red{No pruning:}
    The test accuracy spikes in the early training steps of the globally
    evaluating version. However, the tree quickly overfits and the test
    accuracy of both versions of the algorithm decays to around 60\%.
    This is almost inevitable without pruning for training data with such high
    noise.

  \item \red{Limit on the number of nodes:}
    The most basic idea is to stop training after a certain number of steps,
    i.e., limit the number of nodes.
    We can stop training when the test accuracy spikes, but in practice,
    it is hard to estimate this point.
    The global approach reached a higher peak in \ref{pruning:a}, hence
    seems favorable for this kind of pruning.
    One reason for this is that local evaluation prioritizes splits improving a
    leaf the most, possibly ignoring a split through a leaf that would lead to
    many more instances being classified accurately.
    Another reason is that local evaluation recurses quickly into small
    sub-samples, where the algorithm becomes more sensitively towards noise.

  \item \red{Requiring a minimum of instances in each branch:}
    Other common techniques are requiring a minimum number of instances
    in either leaf nodes ore branches.
    In this example, we require a minimum of 30
    instances in a leaf to perform a further split, giving splits more
    statistic significance.
    We avoid the previous problem of the local evaluation reacting
    sensitive towards noise.
    And in fact, trees with this bias perform better, whereas
    local evaluation outperforms global evaluation.

  \item \red{No distance decay:}
    A parameterless pruning strategy is to stop training once every possible
    split would result in an increased distance.
    This works out well in our example, but there is a risk to already stop
    training in a local minimum.
\end{enumerate}

\noindent
Increasing the bias by pruning only makes sense if the specific bias is actually
apparent in the data.
Otherwise, we decrease the final accuracy of our classifier.
In particular, this means that the success of pruning is highly dependent on the
data and distance measure used.

\begin{figure}[H]
  \begin{subfigure}[t]{0.47\textwidth}
    \includegraphics[width=\textwidth]{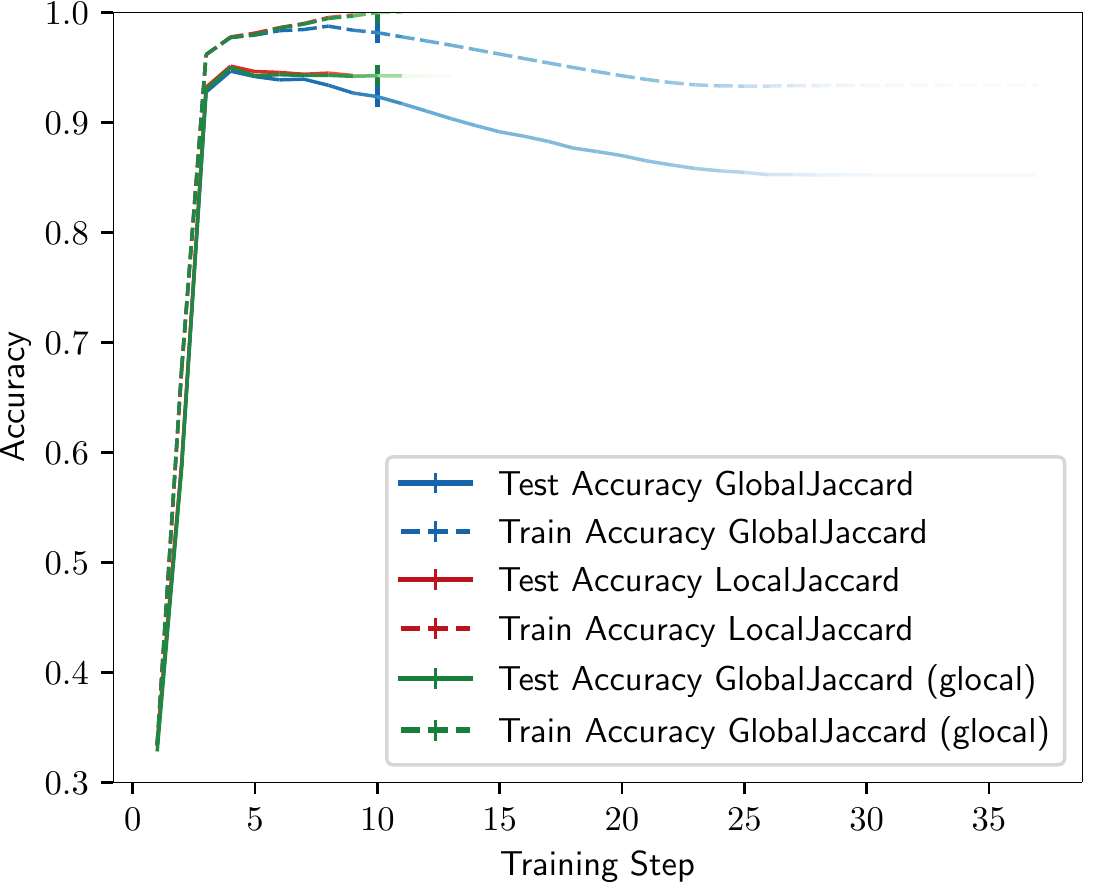}
  \end{subfigure} ~~~~
  \begin{subfigure}[t]{0.47\textwidth}
    \includegraphics[width=\textwidth]{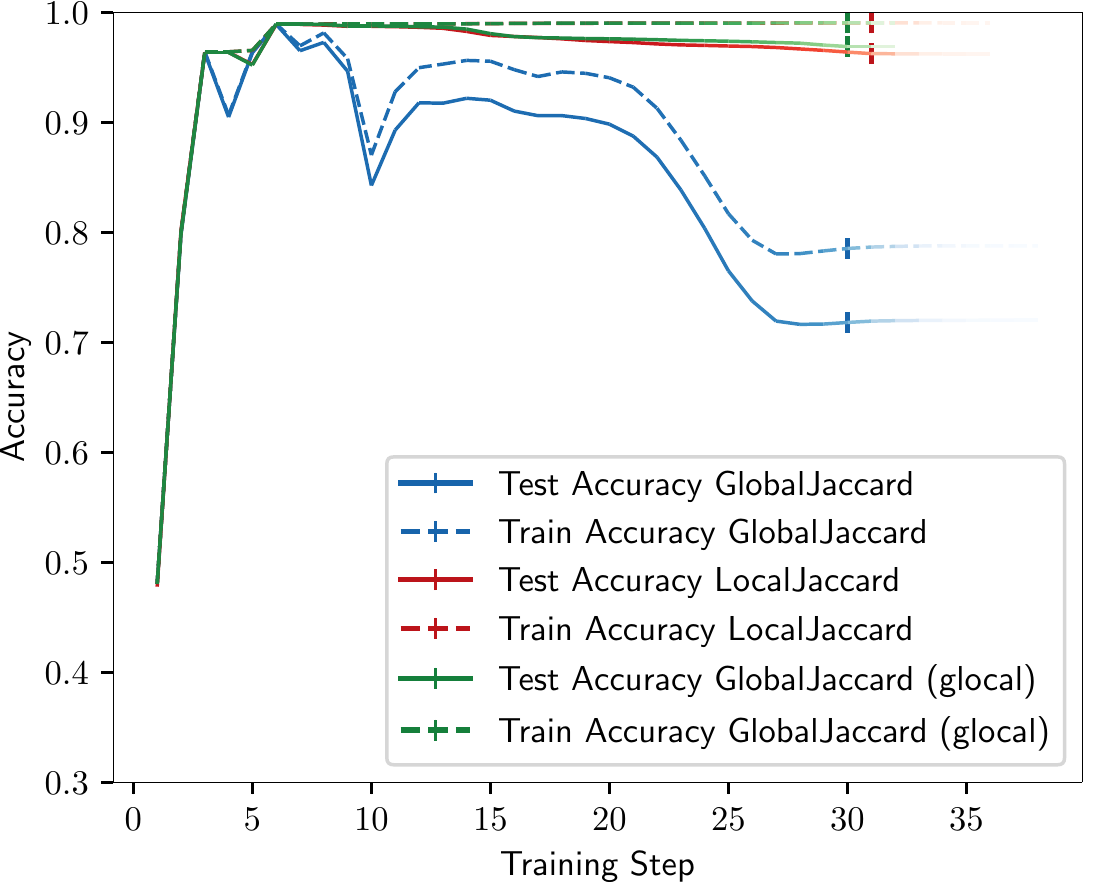}
  \end{subfigure}
  \caption{Train (dashed) and test (solid) accuracy during training of a glocal algorithm.}
  \label{fig:glocal}
\end{figure}

\noindent
One way to make the classifier applicable to a broader
class of datasets is to combine global and local evaluation.
In Figure~\ref{fig:glocal},
we used an approach that works with global evaluation as long as that decreases
the global distance to the ground truth.
Otherwise, a step is performed using the original local approach.
We observe that this helps to overcome the problems we had using the extended Jaccard
distance in the Iris and Monks 3 datasets in Figure~\ref{fig:accts}.

\section{Conclusion}

After introducing multiple distance measures on clusterings and their
application in the ID3 algorithm, we proposed a variation of the algorithm
making decisions based on global distance evaluations.
We evaluated both approaches and found that using local evaluations
generally results in superior accuracy over global evaluations,
which often overgrow trees.
However, the results are not clear cut and we concluded that
the additional bias from larger trees can be useful in scenarios with small
sample size or heavy noise, the latter requiring additional pruning.
Finally, we also saw how the glocal approach alleviates some difficulties with
global evaluation.

We conclude that there is possible application for ID3 under global
evaluations, but the performance depends on the scenario on hand.
Trees created in this fashion can be useful in data analysis by themselves, or
as a random forest, particularly with the shown parameterless pruning strategy.

\bigskip
\noindent
{\em Acknowledgment}: For helpful comments we are grateful to Tobias Sutter (Konstanz).


\begin{small}

\end{small}

\end{document}